\pdfoutput=1

\documentclass[11pt]{article}

\usepackage[final]{acl}
\usepackage{amsmath}
\usepackage{times}
\usepackage{latexsym}
\usepackage{wrapfig}
\usepackage{arydshln}
\usepackage{multirow} 
\usepackage{hyperref}
\usepackage{url}
\usepackage{graphicx}  
\usepackage{booktabs}  
\usepackage{amsthm}
\usepackage{enumerate}
\usepackage{wrapfig}
\usepackage{colortbl}
\usepackage[T1]{fontenc}

\usepackage[utf8]{inputenc}

\usepackage{microtype}

\usepackage{inconsolata}

\usepackage{graphicx}

\title{MoC: Mixtures of Text Chunking Learners for Retrieval-Augmented Generation System}

\author{
  \,Jihao Zhao\textsuperscript{1} \quad
  Zhiyuan Ji\textsuperscript{1} \quad
  Zhaoxin Fan\textsuperscript{2} \quad
  Hanyu Wang\textsuperscript{1} \quad
  Simin Niu\textsuperscript{1} \quad
  Bo Tang\textsuperscript{2} \quad \\
  \,
  \textbf{Feiyu Xiong}\textsuperscript{\textbf{2}} \quad
  \textbf{Zhiyu Li}\textsuperscript{\textbf{2}}\thanks{Corresponding author: \texttt{lizy@iaar.ac.cn}} \\
  \,\textsuperscript{1}School of Information, Renmin University of China, Beijing, China \\
  \,\textsuperscript{2}Institute for Advanced Algorithms Research, Shanghai, China \\
}

\begin{document}
\maketitle
\begin{abstract}
Retrieval-Augmented Generation (RAG), while serving as a viable complement to large language models (LLMs), often overlooks the crucial aspect of text chunking within its pipeline. This paper initially introduces a dual-metric evaluation method, comprising Boundary Clarity and Chunk Stickiness, to enable the direct quantification of chunking quality. Leveraging this assessment method, we highlight the inherent limitations of traditional and semantic chunking in handling complex contextual nuances, thereby substantiating the necessity of integrating LLMs into chunking process. To address the inherent trade-off between computational efficiency and chunking precision in LLM-based approaches, we devise the granularity-aware Mixture-of-Chunkers (MoC) framework, which consists of a three-stage processing mechanism. Notably, our objective is to guide the chunker towards generating a structured list of chunking regular expressions, which are subsequently employed to extract chunks from the original text. Extensive experiments demonstrate that both our proposed metrics and the MoC framework effectively settle challenges of the chunking task, revealing the chunking kernel while enhancing the performance of the RAG system \footnote{Our code is available at \url{https://github.com/IAAR-Shanghai/Meta-Chunking/tree/main/MoC}.}.
\end{abstract}

\section{Introduction}
Retrieval-augmented generation (RAG), as a cutting-edge technological paradigm, aims to address challenges faced by large language models (LLMs), such as data freshness \citep{he2022rethinking}, hallucinations \citep{benedict2023gen,chen2023hallucination,zuccon2023chatgpt,liang2024internal}, and the lack of domain-specific knowledge \citep{li2023chatgpt,shen2023chatgpt}. This is particularly relevant in knowledge-intensive tasks like open-domain question answering (QA) \citep{lazaridou2022internet}. By integrating two key components: the retriever and the generator, this technology enables more precise responses to input queries \citep{singh2021end,lin2023li}. While the feasibility of the retrieval-augmentation strategy has been widely demonstrated through practice, its effectiveness heavily relies on the relevance and accuracy of the retrieved documents \citep{li2022survey,tan2022tegtok}.
The introduction of excessive redundant or incomplete information through retrieval not only fails to enhance the performance of the generation model but may also lead to a decline in answer quality \citep{shi2023large,yan2024corrective}.

In response to the aforementioned challenges, current research efforts mainly focus on two aspects: improving retrieval accuracy \citep{zhuang2024efficientrag,sidiropoulos2022analysing,guo2023prompt} and enhancing the robustness of LLMs against toxic information \citep{longpre2305pretrainer,kim2024adaptive}. However, in RAG systems, a commonly overlooked aspect is the chunked processing of textual content, which directly impacts the quality of dense retrieval for QA \citep{xu2023berm}. This is due to the significant “weakest link” effect in the performance of RAG systems, where the quality of text chunking constrains the retrieved content, thereby influencing the accuracy of generated answers \citep{ru2024ragchecker}. Despite advancements in other algorithmic components, incremental flaws in the chunking strategy can still detract from the overall system performance to some extent.

Given the critical role of text chunking in RAG systems, optimizing this process has emerged as one of the key strategy to mitigate performance bottlenecks. Traditional text chunking methods, often based on rules or semantic similarity \citep{zhang2021sequence,langchain,lyu2024crud}, provide some structural segmentation but are inadequate in capturing subtle changes in logical relationships between sentences. The LumberChunker \citep{duarte2024lumberchunker} offers a novel solution by utilizing LLMs to receive a series of consecutive paragraphs and accurately identify where content begins to diverge. However, it demands a high level of instruction-following ability from LLMs, which incurs significant resource and time costs. Additionally, the effectiveness of current chunking strategies is often evaluated indirectly through downstream tasks, such as the QA accuracy in RAG systems, with a lack of independent metrics for evaluating the inherent rationality of the chunking process itself. These challenges give rise to two practical questions: How can we fully utilize the powerful reasoning capabilities of LLMs while accomplishing the text chunking task at a lower cost? And how to devise evaluation metrics that directly quantify the validity of text chunking?

Inspired by these observations, we innovatively propose two metrics, \textbf{Boundary Clarity} and \textbf{Chunk Stickiness}, to independently and effectively assess chunking quality. Concurrently, we leverage these metrics to delve into the reasons behind the suboptimal performance of semantic chunking in certain scenarios, thereby highlighting the necessity of LLM-based chunking. To mitigate the resource overhead of chunking without compromising the inference performance of LLMs, we introduce the \textbf{Mixture-of-Chunkers (MoC)} framework. This framework primarily comprises a multi-granularity-aware router, specialized meta-chunkers, and a post-processing algorithm. 

This mechanism adopts a divide-and-conquer strategy, partitioning the continuous granularity space into multiple adjacent subdomains, each corresponding to a lightweight, specialized chunker. The router dynamically selects the most appropriate chunker to perform chunking operation based on the current input text. This approach not only effectively addresses the “granularity generalization dilemma” faced by traditional single-model approaches but also maintains computational resource consumption at the level of a single small language model (SLM) through sparse activation, achieving an optimal balance between accuracy and efficiency for the chunking system. It is crucial to emphasize that our objective is not to require the meta-chunker to generate each text chunk in its entirety. Instead, we guide the model to generate a structured list of chunking regular expressions used to extract chunks from the original text. To address potential hallucination phenomena of meta-chunker, we employ an edit distance recovery algorithm, which meticulously compares the generated chunking rules with the original text and subsequently rectifies the generated content.

The main contributions of this work are as follows:
\begin{itemize}
    \item Breaking away from indirect evaluation paradigms, we introduce the dual metrics of Boundary Clarity and Chunk Stickiness to achieve direct quantification of chunking quality. By deconstructing the failure mechanisms of semantic chunking, we provide theoretical validation for the involvement of LLM in chunking tasks.
    \item We devise the MoC architecture, a hybrid framework that dynamically orchestrates lightweight chunking experts via a multi-granularity-aware router. This architecture innovatively integrates: a regex-guided chunking paradigm, a computation resource constraint mechanism based on sparse activation, and a rectification algorithm driven by edit distance.
    \item To validate the effectiveness of our proposed metrics and chunking method, we conduct multidimensional experiments using five different language models across four QA datasets, accompanied by in-depth analysis.
\end{itemize}

\section{Related Works}

\textbf{Text Segmentation} \quad It is a fundamental task in NLP, aimed at breaking down text content into its constituent parts to lay the foundation for subsequent advanced tasks such as information retrieval \citep{li2020neural} and text summarization \citep{lukasik2020text,cho2022toward}. By conducting topic modeling on documents,  \cite{kherwa2020topic} and \cite{barde2017overview} demonstrate the identification of primary and sub-topics within documents as a significant basis for text segmentation. \cite{zhang2021sequence} frames text segmentation as a sentence-level sequence labeling task, utilizing BERT to encode multiple sentences simultaneously. It calculates sentence vectors after modeling longer contextual dependencies and finally predicts whether to perform text segmentation after each sentence. \cite{langchain} provides flexible and powerful support for various text processing scenarios by integrating multiple text segmentation methods, including character segmentation, delimiter-based text segmentation, specific document segmentation, and recursive chunk segmentation. Although these methods better respect the structure of the document, they have limitations in deep contextual understanding. To address this issue, semantic-based segmentation \citep{RetrievalTutorials} utilizes embeddings to aggregate semantically similar text chunks and identifies segmentation points by monitoring significant changes in embedding distances.

\textbf{Text Chunking in RAG}\quad  By expanding the input space of LLMs through introducing retrieved text chunks \citep{guu2020retrieval,lewis2020retrieval}, RAG significantly improves the performance of knowledge-intensive tasks \citep{ram2023context}. Text chunking allows information to be more concentrated, minimizing the interference of irrelevant information, enabling LLMs to focus more on the specific content of each text chunk and generate more precise responses \citep{yu2023chain,besta2024multi,su2024dragin}. LumberChunker \citep{duarte2024lumberchunker} iteratively harnesses LLMs to identify potential segmentation points within a continuous sequence of textual content, showing some potential for LLMs chunking. However, LumberChunker demands a profound capability of LLMs to follow instructions and entails substantial consumption when employing the Gemini model. Notably, access to official API interface of LLM involves cost implications, while common open-source models demonstrate limited generalization capabilities in the domain-specific task of text chunking.

\section{Methodology}
Addressing the current gap in independently evaluating chunking quality, this paper proposes two novel metrics: boundary clarity and chunk stickiness, as detailed in Section \ref{3_1}. Through a preliminary exploration of the chunking mechanism, we elucidate the necessity of leveraging LLMs for chunking tasks and introduce the MoC framework in Section \ref{3_2}. The framework employs a multi-granularity aware routing network with sparse activation to dynamically engage lightweight chunkers. Instead of generating complete text chunks, chunkers are guided to produce structured chunking rule lists, ensuring efficiency optimization without compromising chunking accuracy. A more comprehensive analysis is provided in Appendix \ref{Methodological Analysis}.

\subsection{Deep Reflection on Chunking Strategies}
\label{3_1}
As pointed out by \citet{qu2024semantic}, semantic chunking has not shown a significant advantage in many experiments. This paper further explores this phenomenon and proposes two key metrics to scientifically explain the limitations of semantic chunking and the effectiveness of LLM chunking. At the same time, it also provides independent evaluation indicators for the rationality of chunking itself.

\subsubsection{Boundary Clarity (BC)}
Boundary clarity refers to the effectiveness of chunks in separating semantic units. Specifically, it focuses on whether the structure formed by chunking can create clear boundaries between text units at the semantic level. Blurred chunk boundaries may lead to a decrease in the accuracy of subsequent tasks. Specifically, boundary clarity is calculated utilizing the following formula:
\begin{eqnarray}
\text{BC}(q,d)=\frac{\text{ppl}(q|d)}{\text{ppl}(q)} 
\end{eqnarray}
where $\text{ppl}(q)$ represents the perplexity of sentence sequence $q$, and $\text{ppl}(q|d)$ denotes the \emph{contrastive perplexity} given the text chunk $d$. Perplexity serves as a critical metric for evaluating the predictive accuracy of language models (LMs) on specific textual inputs, where lower perplexity values reveal superior model comprehension of the text, whereas higher values reflect greater uncertainty in semantic interpretation. When the semantic relationship between two text chunks is independent, $\text{ppl}(q|d)$ tends to be closer to $\text{ppl}(q)$, resulting in the BC metric approaching 1. Conversely, strong semantic interdependence drives the BC metric toward zero. Therefore, higher boundary clarity implies that chunks can be effectively separated, whereas a lower boundary clarity indicates blurred boundaries between chunks, which may potentially lead to information confusion and comprehension difficulties.

\begin{figure*}[h]
    \centering
    \includegraphics[width=\textwidth]{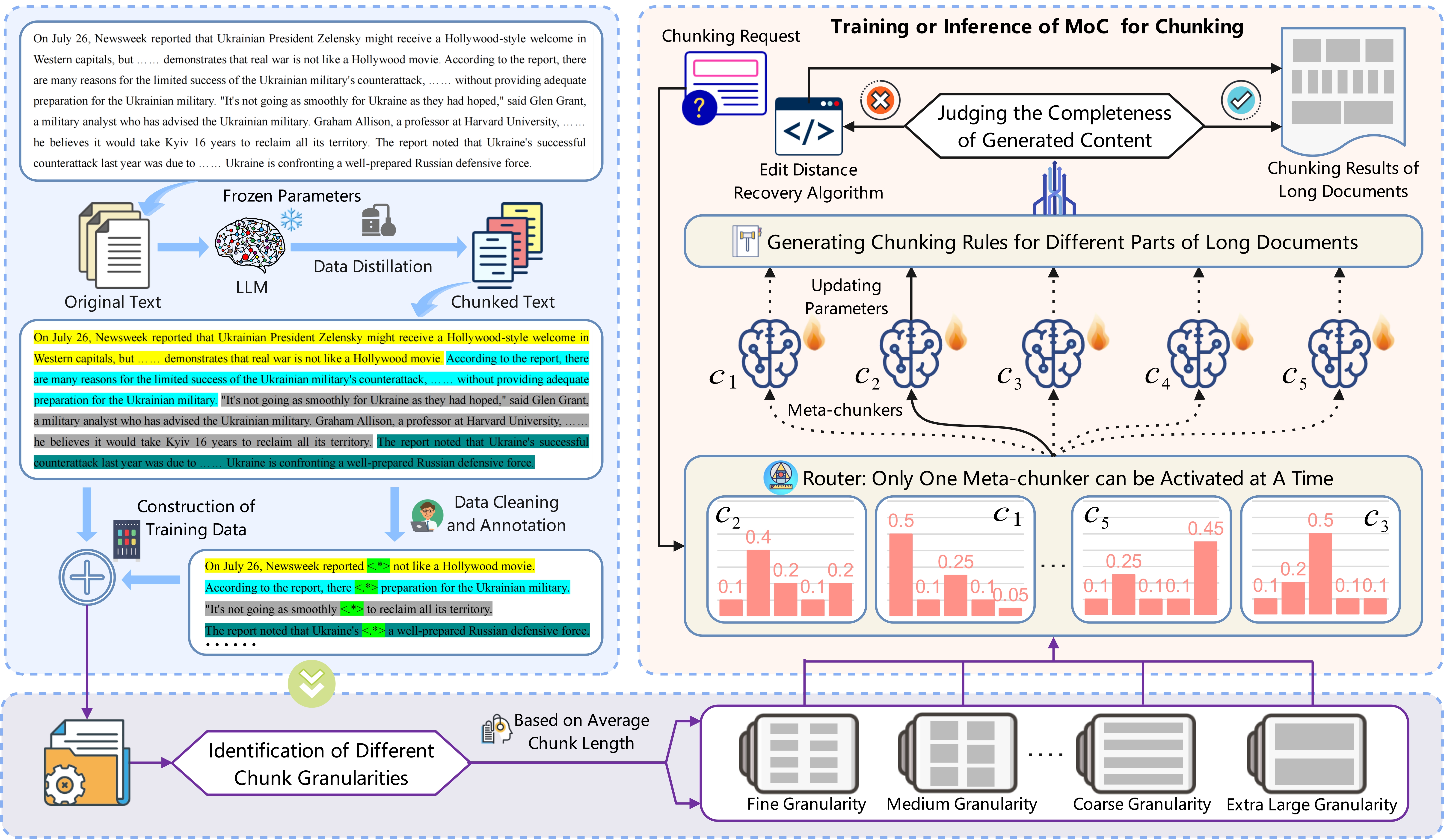}
    \caption{Overview of the entire process of granularity-aware MoC: Dataset construction, training of router and meta-chunkers, as well as chunking inference.}
    \label{fig:rag_pipeline}
\end{figure*}

\subsubsection{Chunk Stickiness (CS)}
The objective of text chunking lies in achieving adaptive partitioning of documents to generate logically coherent independent chunks, ensuring that each segmented chunk encapsulates a complete and self-contained expression of ideas while preventing logical discontinuity during the segmentation process. Chunk stickiness specifically focuses on evaluating the tightness and sequential integrity of semantic relationships between text chunks. This is achieved by constructing a semantic association graph among text chunks, where structural entropy is introduced to quantify the network complexity. Within this graph, nodes represent individual text chunks, and edge weights are defined as follows:
\begin{eqnarray}
\text{Edge}(q,d)=\frac{\text{ppl}(q)-\text{ppl}(q|d)}{\text{ppl}(q)}
\end{eqnarray}
where the theoretical range of the Edge value is defined as $[0,1]$. Specifically, we initially compute the Edge value between any two text chunks within a long document. Values approaching $1$ indicate that $\text{ppl}(q|d)$ tends towards $0$, signifying a high degree of inter-segment correlation. Conversely, an Edge value approaching $0$ suggests that $\text{ppl}(q|d)$ converges to $\text{ppl}(q)$, implying that text chunks are mutually independent. We establish a threshold parameter $K \in (0,1)$ to retain edges exceeding this value. Subsequently, the chunk stickiness is specifically calculated as:
\begin{eqnarray}
\text{CS}(G)=-\sum_{i=1}^{n} \frac{h_i}{2m} \cdot \log_{2}{\frac{h_i}{2m}} 
\end{eqnarray}
where $G$ is the constructed semantic graph, $h_i$ represents the degree of node $i$, and $m$ denotes the total number of edges. This methodology constructs a complete graph, followed by redundancy reduction based on the inter-segment relationships.

On the other hand, to enhance computational efficiency, we construct a sequence-aware incomplete graph that preserves the original ordering of text chunks, which constitutes a graph construction strategy governed by sequential positional constraints. Specifically, given a long text partitioned into an ordered sequence of text chunks $D = \{d_1, d_2, ..., d_n\}$, each node in the graph corresponds to a text chunk, while edge formation is subject to dual criteria: (1) Relevance Criterion: Edge weight $\text{Edge}(d_i,d_j) > K$, where $K$ denotes a predefined threshold; (2) Sequential Constraint: Connections are permitted exclusively when $j - i > \delta$, with $\delta$ representing the sliding window radius fixed at 0. This dual-constraint mechanism strategically incorporates positional relationships, thereby achieving a better equilibrium between semantic relevance and textual coherence.

The detailed design philosophy is elaborated in Appendix \ref{Design Philosophy}. To more intuitively demonstrate the effectiveness of the two metrics, we construct a “Dissimilarity” metric based on the current mainstream semantic similarity, as detailed in Section \ref{Exploring Chunking}. Stemming from the above analysis, we introduce a LM-based training and reasoning framework for text chunking, named granularity-aware MoC.

\subsection{Granularity-Aware MoC}
\label{3_2}
In response to the complex and variable granularity of large-scale text chunking in real-world scenarios, this paper proposes a multi-granularity chunking framework based on MoC. Our approach, whose overall architecture is illustrated in Figure \ref{fig:rag_pipeline}, dynamically routes different granularity experts through a scheduling mechanism and optimizes the integrity of results with a post-processing algorithm.

\subsubsection{Dataset Construction}
We instruct GPT-4o to generate text chunks from raw long-form texts according to the following criteria: (1) Segmentation: The given text should be segmented according to its logical and semantic structure, such that each resulting chunk maintains a complete and independent logical expression. (2) Fidelity: The segmentation outcome must remain faithful to the original text, preserving its vocabulary and content without introducing any fictitious elements. However, extracting such data from GPT-4o poses significant challenges, as the LLM does not always follow instructions, particularly when dealing with long texts that contain numerous special characters. In preliminary experiments, we also observed that GPT-4o tends to alter the expressions used in the original text and, at times, generates fabricated content. 

To address these challenges, we propose the following dataset distillation procedure. We enhance chunking precision in GPT-4o through structured instructions that enforce adherence to predefined rules. A sliding window algorithm, coupled with a chunk buffering mechanism, mitigates the impact of input text length on performance, ensuring seamless transitions between text subsequences. Furthermore, a rigorous data cleaning process, leveraging edit distance calculations and manual review, addresses potential hallucination, while strategic anchor point extraction and placeholder insertion facilitate efficient processing. Detailed implementation and technical specifics are provided in Appendix \ref{Dataset Construction Process}.

\subsubsection{Multi-granularity-aware Router}
\label{Multi-granularity Router}
After the dataset construction is completed, the MoC architecture achieves efficient text processing through the training of the routing decision module and meta-chunkers. The router dynamically evaluates the compatibility of each chunk granularity level based on document features, thereby activating the optimal chunk expert. A major challenge in training the routing module lies in the implicit relationship between text features and chunk granularity, where the goal is to infer the potential granularity of the text without performing explicit chunking operations.

In view of this, we propose a specialized fine-tuning method for SLMs. Firstly, we truncate or concatenate long and short texts respectively, ensuring their lengths hover around 1024 characters. Both operations are performed on text chunks as the operational unit, preserving the semantic integrity of the training texts. By maintaining approximate text lengths, SLMs can better focus on learning features that affect chunk granularity, thus minimizing the impact of text length on route performance. Subsequently, leveraging the segmented data generated by GPT-4o, we assign granularity labels ranging from 0 to 3 to the text, corresponding to average chunk length intervals such as (0, 120], (120, 150], (150, 180], and $(180, +\infty)$. The loss function is formulated as:
\begin{eqnarray}\label{loss function}
\mathcal{L}(\theta) = -\frac{1}{N} \sum_{i=1}^{N} y_i \log(p(y_i|X_i; \theta))
\end{eqnarray}
where $\theta$ represents the set of trainable parameters of the SLM,  $y_i$ denotes the ground-truth granularity label for the $i$-th sample, $N$ signifies the total number of samples, and  $p(y_i|X_i; \theta)$  represents the probability of assigning granularity label $y_i$, given input $X_i$ and current parameters $\theta$.

During inference, we implement marginal sampling over the probability distribution of the final token generated by the SLM in its contextual sequence, selecting the granularity category with the highest probability from the four available categories as the granularity for the corresponding text. Afterwards, the text to be chunked is routed to the corresponding chunking expert:
\begin{eqnarray}
R(X_i) = \arg\max_{k} p(k|X_i; \theta)
\end{eqnarray}
where $k$ represents the category of chunking granularity. Through this mechanism, the router enables dynamic expert selection without explicit chunking operations.

\subsubsection{Meta-chunkers}
\label{Meta-chunkers}
Our objective is not to require meta-chunkers to generate each text chunk in its entirety, but rather to guide it in producing a structured list of segmented regular expressions. Each element in this list contains only the start $S$ and end $E$ of a text chunk $C$, with a special character $r$ replacing the intervening content. The regular expression is represented as:
\begin{eqnarray}
 C_{\text{regex}} = S \oplus r \oplus E, \quad r \in \mathcal{R}
\end{eqnarray}
where $\oplus$ denotes the string concatenation operation, \(\mathcal{R} = \{``<omitted>", ``<ellipsis>", ``[MASK]", ``[ELLIPSIS]", ``.*?", ``<...>", ``<.*>",``<pad>"\}\) is the set of eight special characters we have defined to represent the omitted parts in a text chunk. During the expert training phase, we employ a full fine-tuning strategy, utilizing datasets categorized by different segmentation granularities to optimize the model parameters. The loss function remains consistent with Equation \ref{loss function}. This design allows Meta-chunkers to comprehensively understand the composition of each chunk while significantly reducing the time cost of generation.

\subsubsection{Edit Distance Recovery Algorithm}
\label{Edit Distance}
Let string $A$ denote an element generated by a meta-chunker and string $B$ represent a segment within the original text. The edit distance refers to the minimum number of operations required to transform $A$ into $B$, where the permissible operations include the insertion, deletion, or substitution of a single character. We then define a two-dimensional array, \(\text{ab}[i][j]\), which represents the minimum number of operations needed to convert the substring \(A[1 \ldots i]\) into \(B[1 \ldots j]\). By recursively deriving the state transition formula, we can incrementally construct this array.

Initially, the conditions are as follows: (1) When \(i = 0\), $A$ is an empty string, necessitating the insertion of \(j\) characters to match $B$, thus \(\text{ab}[0][j] = j\); (2) When \(j = 0\), $B$ is an empty string, requiring the deletion of \(i\) characters, hence \(\text{ab}[i][0] = i\); (3) When \(i = j = 0\), the edit distance between two empty strings is evidently \(\text{ab}[0][0] = 0\). Subsequently, the entire \(\text{ab}\) array is populated using the following state transition formula:
\[
\text{ab}[i][j] = 
\begin{cases} 
\text{ab}[i-1][j-1], \quad\quad\   \text{if } A[i] = B[j] \\
1 + \min(\text{ab}[i-1][j], \\ \quad\text{ab}[i][j-1], \\ \quad\text{ab}[i-1][j-1]), \quad \text{if } A[i] \neq B[j]
\end{cases}
\]
If the current characters are identical, no additional operation is required, and the problem reduces to a subproblem; if the characters differ, the operation with the minimal cost among insertion, deletion, or substitution is selected. Ultimately, by utilizing the minimum edit distance, we can accurately pinpoint the field in the original text that most closely matches the elements generated by the meta-chunker, thereby ensuring the precision of regular extraction.

\renewcommand{\arraystretch}{1.4} 
\setlength{\extrarowheight}{1pt} 
\begin{table*}[ht]
\centering
\resizebox{\textwidth}{!}{%
\begin{tabular}{lccc|ccc|c|c}
\toprule
\multirow{2}{*}{\textbf{Chunking Methods}} & \multicolumn{3}{c}{\textbf{CRUD (Single-hop)}} & \multicolumn{3}{c}{\textbf{CRUD (Two-hop)}}  & \multicolumn{1}{c}{\textbf{DuReader}} & \multicolumn{1}{c}{\textbf{WebCPM}} \\
 &\textbf{BLEU-1} & \textbf{BLEU-Avg} & \textbf{ROUGE-L} & \textbf{BLEU-1} & \textbf{BLEU-Avg} & \textbf{ROUGE-L}   & \textbf{F1}  &  \textbf{ROUGE-L} \\
\midrule
Original& 0.3515 & 0.2548 & 0.4213 & 0.2322 & 0.1133 & 0.2613  & 0.2030  & 0.2642\\
Llama\_index& 0.3620 & 0.2682 & 0.4326 & 0.2315 & 0.1133 & 0.2585  & 0.2220 & 0.2630\\
Semantic Chunking& 0.3382 & 0.2462 & 0.4131 & 0.2223 & 0.1075 & 0.2507 & 0.2157 & 0.2691 \\
LumberChunker& 0.3456 & 0.2542 & 0.4160 & 0.2204 & 0.1083 & 0.2521 & 0.2178  & 0.2730\\
\addlinespace[2pt] 
\cdashline{1-9} 
Qwen2.5-14B& 0.3650 & 0.2679 & 0.4351 & 0.2304 & 0.1129 & 0.2587  & 0.2271 & 0.2691\\
Qwen2.5-72B& \underline{0.3721} & \underline{0.2743} & \underline{0.4405} & \textbf{0.2382} & \textbf{0.1185} & \textbf{0.2677}  & \underline{0.2284}  & \underline{0.2693}\\
\addlinespace[2pt] 
\cdashline{1-9} 
Meta-chunker-1.5B& \textbf{0.3754} & \textbf{0.2760} & \textbf{0.4445} & \underline{0.2354} & \underline{0.1155} & \underline{0.2641}  & \textbf{0.2387} & \textbf{0.2745} \\
\bottomrule
\end{tabular}%
}
\caption{Main experimental results are presented in four QA datasets. The best result is in bold, and the second best result is underlined.}
\label{main-performance}
\end{table*}

\section{Experiment}
\subsection{Datasets and Metrics}
We conduct a comprehensive evaluation on three datasets, and covering multiple metrics. The CRUD benchmark \citep{lyu2024crud} contains single-hop and two-hop questions, evaluated using metrics including BLEU series and ROUGE-L. We utilize the DuReader dataset from LongBench benchmark \citep{bai2023longbench}, evaluated based on F1 metric. In addition, a dataset called WebCPM \citep{qin2023webcpm} specifically designed for long-text QA, is utilized to retrieve relevant facts and generate detailed paragraph-style responses, with ROUGE-L as the metric.

\subsection{Baselines}
We primarily compare meta-chunker and MoC with two types of baselines, namely rule-based chunking and dynamic chunking, noting that the latter incorporates both semantic similarity models and LLMs. The original rule-based method simply divides long texts into fixed-length chunks, disregarding sentence boundaries. However, the Llama\_index method \citep{langchain} offers a more nuanced approach, balancing the maintenance of sentence boundaries while ensuring that token counts in each segment are close to a preset threshold. On the other hand, semantic chunking \citep{xiao2023c} utilizes sentence embedding models to segment text based on semantic similarity. LumberChunker \citep{duarte2024lumberchunker} employs LLMs to predict optimal segmentation points within the text.

\subsection{Experimental Settings}
Without additional annotations, all LMs used in this paper adopt chat or instruction versions. When chunking, we primarily employ LMs with the following hyperparameter settings: temperature at 0.1 and top-p at 0.1. For evaluation, Qwen2-7B is applied with the following settings: top\_p = 0.9, top\_k = 5, temperature = 0.1, and max\_new\_tokens = 1280. When conducting QA, the system necessitates dense retrievals from the vector database, with top\_k set to 8 for CRUD, 5 for DuReader and WebCPM. To control variables, we maintain a consistent average chunk length of 178 for various chunking methods across each dataset. Detailed experimental setup information can be found in Appendix \ref{appendix:1}.

\subsection{Main Results}
\label{sec4_1:Main_Results}
To comprehensively validate the effectiveness of the proposed meta-chunker and MoC architectures, we conducts experiments using three widely used QA datasets. During dataset preparation, we curate 20,000 chunked QA pairs through rigorous processing. Initially, we fine-tune the Qwen2.5-1.5B model using this data. As shown in Table \ref{main-performance}, compared to traditional rule-based and semantic chunking methods, as well as the state-of-the-art LumberChunker approach based on Qwen2.5-14B, the Meta-chunker-1.5B exhibits both improved and more stable performance. Furthermore, we directly perform chunking employing Qwen2.5-14B and Qwen2.5-72B. The results demonstrate that these LLMs, with their powerful context processing and reasoning abilities, also deliver outstanding performance in chunking tasks. However, Meta-chunker-1.5B slightly underperforms the 72B model only in the two-hop CRUD, while outperforming both LLMs in other scenarios.

\renewcommand{\arraystretch}{1.4} 
\setlength{\extrarowheight}{1pt} 
\begin{table}[t]
\centering
\resizebox{0.47\textwidth}{!}{%
\begin{tabular}{lccccc}
\toprule
 \textbf{Methods} & \textbf{BLEU-1} & \textbf{BLEU-2} & \textbf{BLEU-3} & \textbf{BLEU-4} & \textbf{ROUGE-L}   \\
\midrule
\text{<pad>} & 0.3683 & 0.2953 & 0.2490 & 0.2132 & 0.4391 \\
\text{<omitted>} & 0.3725 & 0.2985 & 0.2523 & 0.2165 & 0.4401   \\
\text{<ellipsis>} & 0.3761 & 0.3025 & 0.2554 & 0.2193 & 0.4452   \\
\text{[MASK]} & 0.3754 & 0.3012 & 0.2545 & 0.2188 & 0.4445 \\
\text{[ELLIPSIS]} & 0.3699 & 0.2966 & 0.2510 & 0.2159 & 0.4380 \\
\text{.*?} & 0.3745 & 0.3015 & 0.2553 & 0.2195 & 0.4437 \\
\text{<...>} & 0.3716 & 0.2988 & 0.2526 & 0.2167 & 0.4412 \\
\text{<.*>} & 0.3790 & 0.3054 & 0.2583 & 0.2221 & 0.4470 \\
\addlinespace[2pt] 
\cdashline{1-6} 
MoC & \textbf{0.3826} & \textbf{0.3077} & \textbf{0.2602} & \textbf{0.2234} & \textbf{0.4510} \\
\bottomrule
\end{tabular}%
}
\caption{Performance impact of special characters and the effectiveness of granularity-aware  MoC framework in text chunking.}
\label{tab:3}
\end{table}

Upon validating the effectiveness of our proposed chunking experts, we proceeded to investigate the impact of various special characters on performance, and extended chunking within the MoC framework. As illustrated in Table \ref{tab:3}, we design eight distinct special characters, each inducing varying degrees of performance fluctuation in the meta-chunker. Notably, all character configurations demonstrate measurable performance enhancements compared to baseline approaches, with $[Mask]$ and $<.*>$ exhibiting particularly remarkable efficacy. In our experiments, both the Meta-chunker-1.5B and the MoC framework employ $[Mask]$ as an ellipsis to replace the middle sections of text chunks, while maintaining consistent training data. The experimental results indicate that the chunking method based on the MoC architecture further enhances performance. Specifically, when handling complex long texts, MoC effectively differentiates the chunking granularity of various sections. Moreover, the time complexity of the MoC remains at the level of a single SLM, showcasing a commendable balance between computational efficiency and performance. Further details regarding analysis and experiments are outlined in Appendices \ref{Another Perspective} and \ref{Chunking Efficiency}.

\renewcommand{\arraystretch}{1.4} 
\setlength{\extrarowheight}{1pt} 
\begin{table*}[h]
\centering
\resizebox{\textwidth}{!}{%
\begin{tabular}{lccc|ccc|ccc|ccc}
\toprule
\multirow{2}{*}{\textbf{Chunking Methods}} & \multicolumn{3}{c}{\textbf{Qwen2.5-1.5B}} & \multicolumn{3}{c}{\textbf{Qwen2.5-7B}} & \multicolumn{3}{c}{\textbf{Qwen2.5-14B}} & \multicolumn{3}{c}{\textbf{Internlm3-8B}}  \\
 &\textbf{BC} & \textbf{$\text{CS}_c$} & \textbf{$\text{CS}_i$} & \textbf{BC} & \textbf{$\text{CS}_c$} & \textbf{$\text{CS}_i$} &\textbf{BC} & \textbf{$\text{CS}_c$} & \textbf{$\text{CS}_i$} &\textbf{BC} & \textbf{$\text{CS}_c$} & \textbf{$\text{CS}_i$}   \\
\midrule
Original & 0.8210 & 2.397 & 1.800 & 0.8049 & 2.421 & 1.898 & 0.7704 & 2.297 & 1.459 & 0.8054 & 2.409 & 1.940 \\
Llama\_index& 0.8590 & 2.185 & 1.379 & 0.8455 & 2.250 & 1.483 & 0.8117 & 2.081 & 1.088 & 0.8334 & 2.107 & 1.303 \\
Semantic Chunking& 0.8260 & 2.280 & 1.552 & 0.8140 & 2.325 & 1.650 & 0.7751 & 2.207 & 1.314 & 0.8027 & 2.255 & 1.546 \\
Qwen2.5-14B & \textbf{0.8750} & \textbf{2.069} & \textbf{1.340} & \textbf{0.8641} & \textbf{2.125} & \textbf{1.438} & \textbf{0.8302} & \textbf{1.927} & \textbf{1.068} & \textbf{0.8444} & \textbf{1.889} & \textbf{1.181} \\
\bottomrule
\end{tabular}%
}
\caption{Performance of different chunking methods under various LMs, directly calculated using two metrics we proposed: BC represents boundary clarity, which is preferable when higher; $\text{CS}_c$ denotes chunk stickiness utilizing a complete graph, and $\text{CS}_i$ indicates chunk stickiness employing a incomplete graph, both of which are favorable when lower.}
\label{two metrics we proposed}
\end{table*}

\subsection{Exploring Chunking Based on Boundary Clarity and Chunk Stickiness}
\label{Exploring Chunking}
To compare the effectiveness of the two metrics we designed, we introduce the "Dissimilarity" (DS) metric: 
\begin{eqnarray*}
\text{DS} = 1 - \text{sim}(q, d)
\end{eqnarray*}
where \(\text{sim}(q, d)\) represents the semantic similarity score between the text chunks \(q\) and \(d\). With this definition, the DS metric ranges from [0, 1], where 0 indicates perfect similarity and 1 indicates complete dissimilarity. The design of the DS metric is based on the following considerations: first, semantic similarity measures are typically employed to assess the degree of semantic proximity between two text segments. By converting this to the dissimilarity measure, we can more directly observe the semantic differences between chunks. Second, the linear transformation of DS preserves the monotonicity of the original similarity measure without losing any information. The left side of Table \ref{tab:DS} reveals the QA performance of RAG using different chunking strategies. It is important to note that, to ensure the validity of the evaluation, we maintained the same average text chunk length across all chunking methods.

\renewcommand{\arraystretch}{1.4} 
\setlength{\extrarowheight}{1pt} 
\begin{table}[h]
\centering
\resizebox{0.47\textwidth}{!}{%
\begin{tabular}{lcc|cc}
\toprule
\multirow{2}{*}{\textbf{Chunking Methods}} & \multicolumn{2}{c}{\textbf{QA Performance}} & \multicolumn{2}{c}{\textbf{Dissimilarity}}   \\
 &\textbf{BLEU-1} & \textbf{ROUGE-L} & \textbf{Model-1} & \textbf{Model-2}  \\
\midrule
Original & 0.3515 & 0.4213 & 0.2731 & 0.2885 \\
Llama\_index& 0.3620 & 0.4326 & 0.2836 & 0.3071 \\
Semantic Chunking& 0.3382 & 0.4131 & \textbf{0.4174} & \textbf{0.4162} \\
Qwen2.5-14B & \textbf{0.3650} & \textbf{0.4351} & 0.2955 & 0.3263 \\
\bottomrule
\end{tabular}%
}
\caption{QA performance across different chunking methods and the degree of dissimilarity between text chunks, with Model-1 and Model-2 denoting semantic similarity models bge-large-zh-v1.5 and all-MiniLM-L6-v2, Respectively.}
\label{tab:DS}
\end{table}

\textbf{Why Does Semantic Chunking Underperform?}\quad 
As illustrated on the right side of Table \ref{tab:DS}, while semantic chunking scores are generally high, its performance in QA tasks is suboptimal. Moreover, there is no evident correlation between the scores of semantic dissimilarity and the efficacy of QA. This suggests that in the context of RAG, relying solely on semantic similarity between sentences is insufficient for accurately delineating the optimal boundaries of text chunks. For further in-depth analyses and illustrations, refer to Appendix \ref{Relationship3} and Figure \ref{picExploring Chunking}.

Furthermore, it can be observed from Table \ref{two metrics we proposed} that the clarity of semantic chunking boundaries is only marginally superior to fixed-length chunking. This implies that although semantic chunking attempts to account for the degree of association between sentences, its limited ability to distinguish logically connected sentences often results in incorrect segmentation of content that should remain coherent. Additionally, Table \ref{two metrics we proposed} reveals that semantic chunking also falls short in terms of capturing semantic relationships, leading to higher chunk stickiness and consequently affecting the independence of text chunks.

\textbf{Why Does LLM-Based Chunking Work?}\quad 
As shown in Table \ref{two metrics we proposed}, text chunks generated by LLMs exhibit superior boundary clarity, indicating the heightened ability to accurately identify semantic shifts and topic transitions, thereby mitigating the erroneous segmentation of related sentences. Concurrently, the LLM-based chunking produces chunks with reduced chunk stickiness, signifying that the internal semantics of chunks are more tightly bound, while a greater degree of independence is maintained between chunks. We also validate that BC and CS exhibit a significant correlation with the QA performance, as detailed in Appendix \ref{Relationship3}. Accordingly, this combination of well-defined boundaries and diminished stickiness contributes to enhanced retrieval efficiency and generation quality within RAG systems, ultimately leading to superior overall performance.

\subsection{Hyper-parameter Sensitivity Analysis}
In calculating the chunk stickiness, we rely on the \( K \) to filter out edges with weaker associations between text chunks in the knowledge graph. As presented in Table \ref{tab:6}, an increase in the value of \( K \) leads to a gradual decrease in the metric. This occurs because a larger \( K \) value limits the number of retained edges, resulting in a sparser connectivity structure within the graph. Notably, regardless of the chosen \( K \) value, the LLM-based chunking method consistently maintains a low level of chunk stickiness. This indicates that it more accurately identifies semantic transition points between sentences, effectively avoiding excessive cohesion between text chunks caused by interruptions within paragraphs. 

\renewcommand{\arraystretch}{1.4} 
\setlength{\extrarowheight}{1pt} 
\begin{table}[h]
\centering
\resizebox{0.47\textwidth}{!}{%
\begin{tabular}{lccc|ccc}
\toprule
\multirow{2}{*}{\textbf{Chunking Methods}} & \multicolumn{3}{c}{\textbf{Complete Graph}} & \multicolumn{3}{c}{\textbf{Incomplete Graph}}   \\
 &\textbf{0.7} & \textbf{0.8} & \textbf{0.9} & \textbf{0.7} & \textbf{0.8} & \textbf{0.9}  \\
\midrule
Original& 2.536& 2.397& 2.035& 2.199& 1.800& 1.300 \\
Llama\_index& 2.454& 2.185& 1.543& 1.997& 1.379& 0.740 \\
Semantic Chunking& 2.455& 2.280& 1.733& 2.039& 1.552& 0.835 \\
Qwen2.5-14B &\textbf{2.364}& \textbf{2.069}& \textbf{1.381}& \textbf{1.972}& \textbf{1.340}& \textbf{0.623} \\
\bottomrule
\end{tabular}%
}
\caption{Performance sensitivity of $K$ in chunk stickiness.}
\label{tab:6}
\end{table}

\begin{figure}[h]
    \centering
    \includegraphics[width=0.47\textwidth]{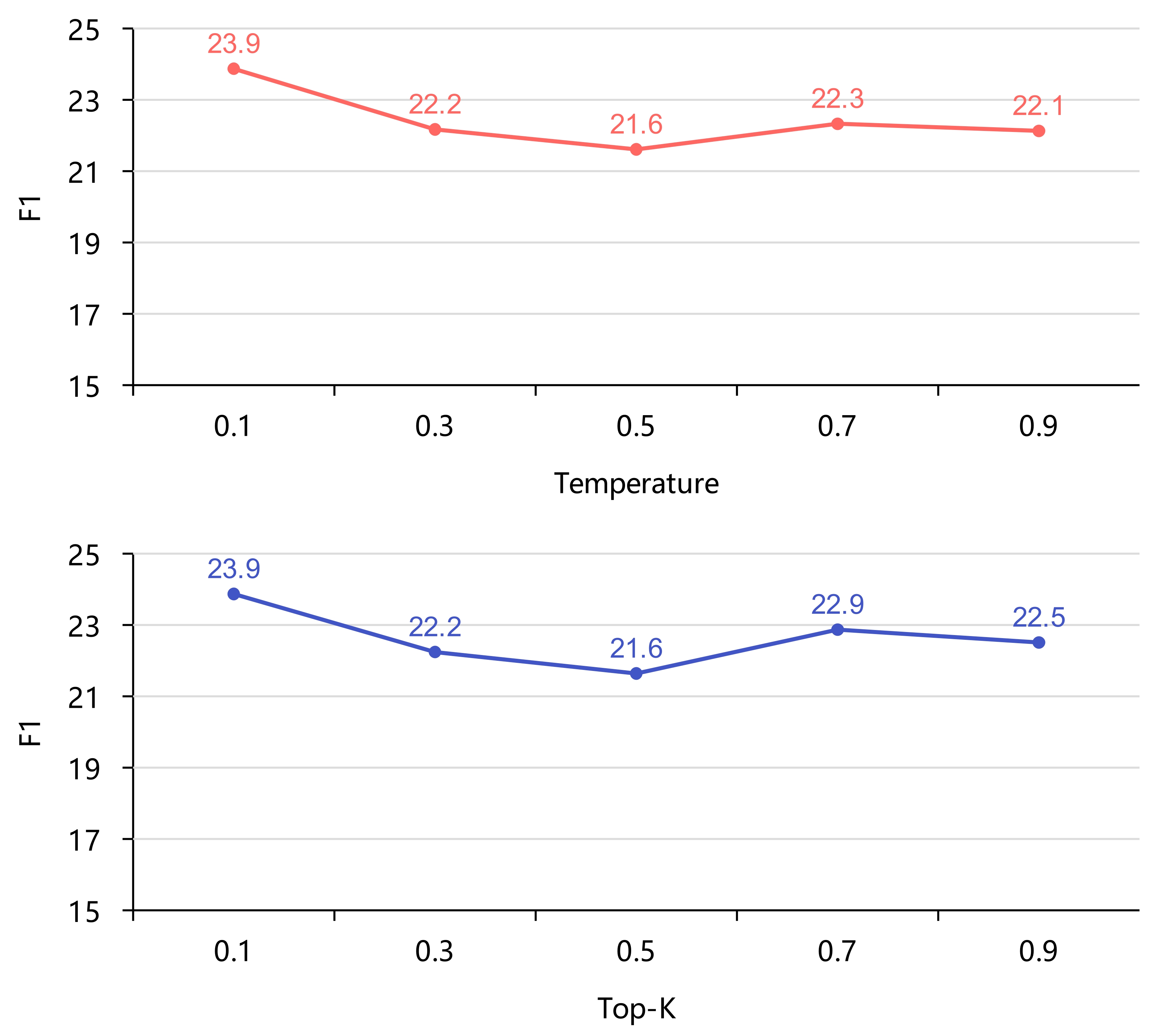}
    \caption{Performance sensitivity to temperature and top-k.}
    \label{biaoSensitivity Analysis}
\end{figure}

We conduct experiments on the decoding sampling hyperparameters of the meta-chunker within the MoC framework, with specific results presented in Figure \ref{biaoSensitivity Analysis}. Experimental data demonstrates that higher values of temperature and top-k sampling strategies introduce increased randomness, thereby exerting a certain impact on the chunking effect. Conversely, when these two hyperparameters are set to lower values, the model typically provides more stable and precise chunking, leading to a more significant performance improvement.

\section{Conclusion}
Addressing the current void in the independent assessment of chunking quality, this paper introduces two novel evaluation metrics: boundary clarity and chunk stickiness.  It systematically elucidates the inherent limitations of semantic chunking in long-text processing, which further leads to the necessity of LLM-based chunking. Amidst the drive for performance and efficiency optimization, we propose the MoC framework, which utilizes sparsely activated meta-chunkers through multi-granularity-aware router. It's worth emphasizing that this study guides meta-chunkers to generate a highly structured list of chunking regular expressions, precisely extracting text chunks from the original text using only a few characters from the beginning and end. Our approach demonstrates superior performance compared to strong baselines.

\section{Limitations}
Despite the superior performance demonstrated by the proposed MoC framework for chunking tasks on various datasets, there are still some limitations that merit further exploration and improvement. Although we have implemented multiple quality control measures to ensure data quality and constructed a training set consisting of nearly 20,000 data entries, the current dataset size remains relatively limited compared to the massive scale and complex diversity of real-world text data. We have mobilized the power of the open-source community to further enrich our chunking dataset utilizing pre-training data from LMs. Additionally, while the dataset construction process is flexible and theoretically expandable to more scenarios, it has not yet undergone adequate multi-language adaptation and validation. We leave this aspect for future research.

\bibliography{custom}

\appendix
\section{Appendix}
\subsection{Main Experimental Details}
\label{appendix:1}
All language models utilized in this paper employ the chat or instruct versions where multiple versions exist, and are loaded in full precision. The vector database is constructed using Milvus, where the embedding model is bge-large-zh-v1.5. In experiments, we utilize a total of four benchmarks, and their specific configurations are detailed as follows:

\begin{enumerate}[(a)]
    \item \textbf{Rule-based Chunking Methods}
    \begin{itemize}
        \item \textbf{Original}: This method divides long texts into segments of a fixed length, such as two hundred Chinese characters or words, without considering sentence boundaries.
        
        \item \textbf{Llama\_index} \citep{langchain}: This method considers both sentence completeness and token counts during segmentation. It prioritizes maintaining sentence boundaries while ensuring that the number of tokens in each chunk are close to a preset threshold. We use the \texttt{SimpleNodeParser} function from \texttt{Llama\_index}, adjusting the \texttt{chunk\_size} parameter to control segment length. Overlaps are handled by dynamically overlapping segments using the \texttt{chunk\_overlap} parameter, ensuring sentence completeness during segmentation and overlapping.
    \end{itemize}
    
    \item \textbf{Dynamic Chunking Methods}
    \begin{itemize}
        \item \textbf{Semantic Chunking} \citep{xiao2023c}: Utilizes pre-trained sentence embedding models to calculate the cosine similarity between sentences. By setting a similarity threshold, sentences with lower similarity are selected as segmentation points, ensuring that sentences within each chunk are highly semantically related. This method employs the \texttt{SemanticSplitterNodeParser} from \texttt{Llama\_index}, exploiting the bge-base-zh-v1.5 model. The size of the text chunks is controlled by adjusting the similarity threshold.
        
        \item \textbf{LumberChunker} \citep{duarte2024lumberchunker}: Leverages the reasoning capabilities of LLMs to predict suitable segmentation points within the text. We utilize Qwen2.5 models with 14B parameters, set to full precision.
    \end{itemize}
\end{enumerate}

\subsection{Design Philosophy of Chunk Stickiness}
\label{Design Philosophy}
In the context of network architecture, high structural entropy tends to exhibit greater challenges in predictability and controllability due to its inherent randomness and complexity. Our chunking strategy aims to maximize semantic independence between text chunks while maintaining a coherent semantic expression. Consequently, a higher chunk stickiness implies greater interconnectedness among these chunks, resulting in a more intricate and less ordered semantic network. Furthermore, to ensure a robust comparison between different chunking methods, we enforce a uniform average chunking length. This standardization provides a fair basis for evaluation, mitigating potential biases arising from discrepancies in chunking size. Ultimately, a lower CS score signifies that the chunking method is more accurate in identifying semantic transition points between sentences, thereby avoiding the fragmentation of coherent passages and the consequent excessive stickiness between resulting chunks.

To more intuitively demonstrate the effectiveness of the two metrics we designed, we construct a “Dissimilarity” metric based on the current mainstream semantic similarity, as detailed in Section \ref{Exploring Chunking}. Furthermore, employing several chunking techniques and LLMs, we conduct an in-depth investigation of boundary clarity and chunk stickiness, conducting comparative experiments with the dissimilarity metric. The experimental results clearly show that the two proposed metrics exhibit a consistent trend with RAG performance when evaluating the quality of text chunking. In contrast, the dissimilarity metric fail to display a similar variation. This suggests that, even without relying on QA accuracy, the two proposed metrics can independently and effectively assess chunking quality.

\subsection{Dataset Construction Process}
\label{Dataset Construction Process}
\textbf{Structured Instruction Design}\quad By explicitly enumerating rules, GPT-4o is compelled to adhere to predefined chunking regulations, such as ensuring semantic unit integrity, enforcing punctuation boundaries, and prohibiting content rewriting. 

\textbf{Sliding Window and Chunk Buffering Mechanism}\quad Drawing from the research  conducted by \citet{duarte2024lumberchunker} and practical experience, we observe that the length of the original text significantly influences the chunking performance of LLMs. To address this problem, we initially apply a sliding window algorithm to segment the input text into subsequences, each below a threshold of 1024 tokens. Segmentation points are prioritized at paragraph boundaries or sentence-ending positions. These subsequences are then processed sequentially by GPT-4o. To maintain continuity between two consecutive subsequences, we implement a chunk buffer mechanism by removing the last generated text chunk of the preceding sequence and using it as the prefix for the subsequent sequence, thereby ensuring smooth information flow.

\textbf{Data Cleaning and Annotation}\quad To identify and eliminate hallucinated content during the generation process, we calculate the difference between each chunk and the paragraphs in the original text through the edit distance, as outlined in Section \ref{Edit Distance}. If the minimum edit distance exceeds 10\% of the chunk length, we manually review the location of the chunk error and make corrections accordingly. Additionally, for a long text, we extract several characters at the beginning and end of each text chunk as anchor points, while replacing the intermediate content with eight preset special placeholders, as demonstrated in Sections \ref{Multi-granularity Router} and \ref{Meta-chunkers}.

\textbf{Classification of Granularity Labels} The division of granularity labels is strictly based on the statistical properties of the data distribution. Granularity labels are used to partition the original data into independent training sets, with a balanced amount of data corresponding to each label. Specifically, the total amount of data we processed is 20K. To ensure sufficient data under each label for adequate training of each chunker, we strive to maintain approximately 5K data points per label. In terms of independence, we ensure that each granularity label is relatively independent, meaning there is no overlap or significant correlation between labels, thus preventing confusion during the training process of chunkers. 

Moreover, given the relative scarcity of datasets for text chunking, we have expanded our chunking dataset to 120K entries, with 60K entries each in Chinese and English, and have made it open-source for sharing within the academia. We are also continuously optimizing the dataset with the aim of further improving its quality. This will provide a more solid foundation for the division of granularity labels, ensuring that each chunker can be adequately trained with sufficient data support, thereby possessing greater robustness.

\subsection{Another Perspective on Chunking Performance Comparison}
\label{Another Perspective}
The performance evaluation of RAG systems primarily focuses on the similarity between generated answers and reference answers. However, this evaluation method introduces additional noise during the decoding strategy in the generation stage, making it difficult to distinguish whether the performance defects originate from the retrieved chunk or the generation module. To address this constraint, we propose an evaluation approach based on information support, which centers on quantifying the supporting capability of retrieved text chunks for the target answer through conditional probability modeling.

Given a set of retrieved chunks \( C = \{c_1, c_2, ..., c_n\} \) and the reference answer \( A = \{a_1, a_2, ..., a_m\} \), we employ a LLM to compute the average conditional probability (CP) of the target answer:
\begin{eqnarray}
\text{CP}=-\frac{1}{M} \sum_{i=1}^{M} \log P(a_i | c_1, c_2, \ldots, c_{n})
\end{eqnarray}
A smaller CP value indicates a higher likelihood of the correct answer being inferred from the retrieved text chunks, signifying stronger support. The results presented in Table \ref{tab:4} show that, even when evaluated with different LMs, our chunking method consistently exhibits high support. This suggests that our chunking strategy, by optimizing the semantic integrity and independence of text chunks, enhances the relevance of the retrieved text to the question, thereby reducing the difficulty of generating the correct answer.

\renewcommand{\arraystretch}{1.4} 
\setlength{\extrarowheight}{1pt} 
\begin{table}[t]
\centering
\resizebox{0.47\textwidth}{!}{%
\begin{tabular}{lcccc}
\toprule
 \textbf{Chunking Methods} & \textbf{Qwen2.5-1.5B} & \textbf{Qwen2.5-7B} & \textbf{Qwen2.5-14B} & \textbf{Internlm3-8B}   \\
\midrule
Original & 2.206 & 2.650 & 2.560 & 1.636 \\
Llama\_index & 1.964 & 2.412 & 2.353 & 1.486 \\
Semantic Chunking & 1.865 & 2.331 & 2.238 & 1.411 \\
LumberChunker & 2.184 & 2.593 & 2.589 & 1.652 \\
Qwen2.5-14B & 1.841 & 2.313 & 2.209 & 1.373 \\
Meta-chunker-1.5B & \textbf{1.835} & \textbf{2.267} & \textbf{2.199} & \textbf{1.367} \\
\bottomrule
\end{tabular}%
}
\caption{Information-based performance evaluation for the RAG system.}
\label{tab:4}
\end{table}

\subsection{Exploration of Text Chunking Efficiency}
\label{Chunking Efficiency}
To address the issue of complex and variable text chunking granularities in real-world scenarios, this paper proposes the multi-granularity chunking framework. This method schedules chunkers for different granularities through a dynamic routing mechanism. This sparse activation mechanism controls computational resource consumption at the level of a single chunker, enabling the overall system to optimize efficiency while maintaining accuracy.

\begin{itemize}
    \item This mechanism effectively resolves the granularity generalization dilemma faced by traditional single models, as shown in Figure \ref{Chunk_granularity}: when training data covers a wide range of chunking granularities, SLMs with limited parameters struggle to capture multi-granularity feature correlations, while simply scaling up the model size leads to exponential growth in computational complexity. 
    \item The MoC framework decouples complex multi-granularity chunking tasks into multiple expert subtasks, allowing each chunker to focus solely on feature modeling within a specific granularity range. 
    \item It is worth emphasizing that our goal is not to require the model to generate every text chunk in its entirety, but rather to guide it in generating a structured list of chunking regular expressions. This approach offers dual advantages: on the one hand, it fully leverages the powerful reasoning capabilities of language models to ensure the reasonability of the chunking; on the other hand, by eliminating the need to generate the complete content of each text chunk, it saves generation time and improves overall processing efficiency.
\end{itemize}

To quantify the speed of chunking, we evaluate various chunking methods, which necessitate the utilization of models, using a single NVIDIA A800 GPU on the CRUD benchmark. The experimental results demonstrate that, although the MoC design incurs some additional computational overhead, we are able to effectively control the processing time while maintaining accuracy through the sparse activation mechanism and the use of 1.5B chunkers, as shown in Table \ref{tab:Average Time}. Furthermore, in RAG scenarios, chunking operations are usually used to construct a local vector knowledge base, which can be reused multiple times. Consequently, the emphasis is placed more on the precision of chunking in applications.

\renewcommand{\arraystretch}{1.4} 
\setlength{\extrarowheight}{1pt} 
\begin{table}[h]
\centering
\resizebox{0.47\textwidth}{!}{%
\begin{tabular}{lc}
\toprule
 \textbf{Chunking Methods} & \textbf{Average Time Per Document (Unit: s)}    \\
\midrule
Semantic Chunking & 1.16  \\
LumberChunker & 3.23  \\
Qwen2.5-14B & 26.99  \\
Meta-chunker-1.5B & 3.69  \\
\bottomrule
\end{tabular}%
}
\caption{Efficiency comparison of different model-based chunking methods on the CRUD benchmark, with the unit defined as the average processing time in seconds per document.}
\label{tab:Average Time}
\end{table}

\subsection{Methodological Analysis}
\label{Methodological Analysis}
We firstly reorganize and briefly introduce the framework of the paper to convey our research ideas more clearly. This paper initially employs structured instruction design to explicitly guide the GPT-4o model\footnote{\url{https://openai.com/api/}} in text chunking based on logical and semantic structures. In this process, we innovatively adopt a sliding window algorithm and a chunk buffering mechanism to handle long texts, ensuring information continuity. Simultaneously, to directly measure the quality of chunking, we propose two new metrics, boundary clarity and chunk stickiness, to strictly control the chunking results. Subsequently, when constructing the training dataset, we extract the first and last characters of text chunks as anchors, with the intermediate content replaced by special placeholders, thereby generating high-quality training samples. Our objective is to guide a SLM to generate a structured list of chunking regular expressions and then extract text chunks from the original text, resolving the conflict between LLM computational efficiency and chunking accuracy.

\begin{figure}[h]
    \centering
    \includegraphics[width=0.47\textwidth]{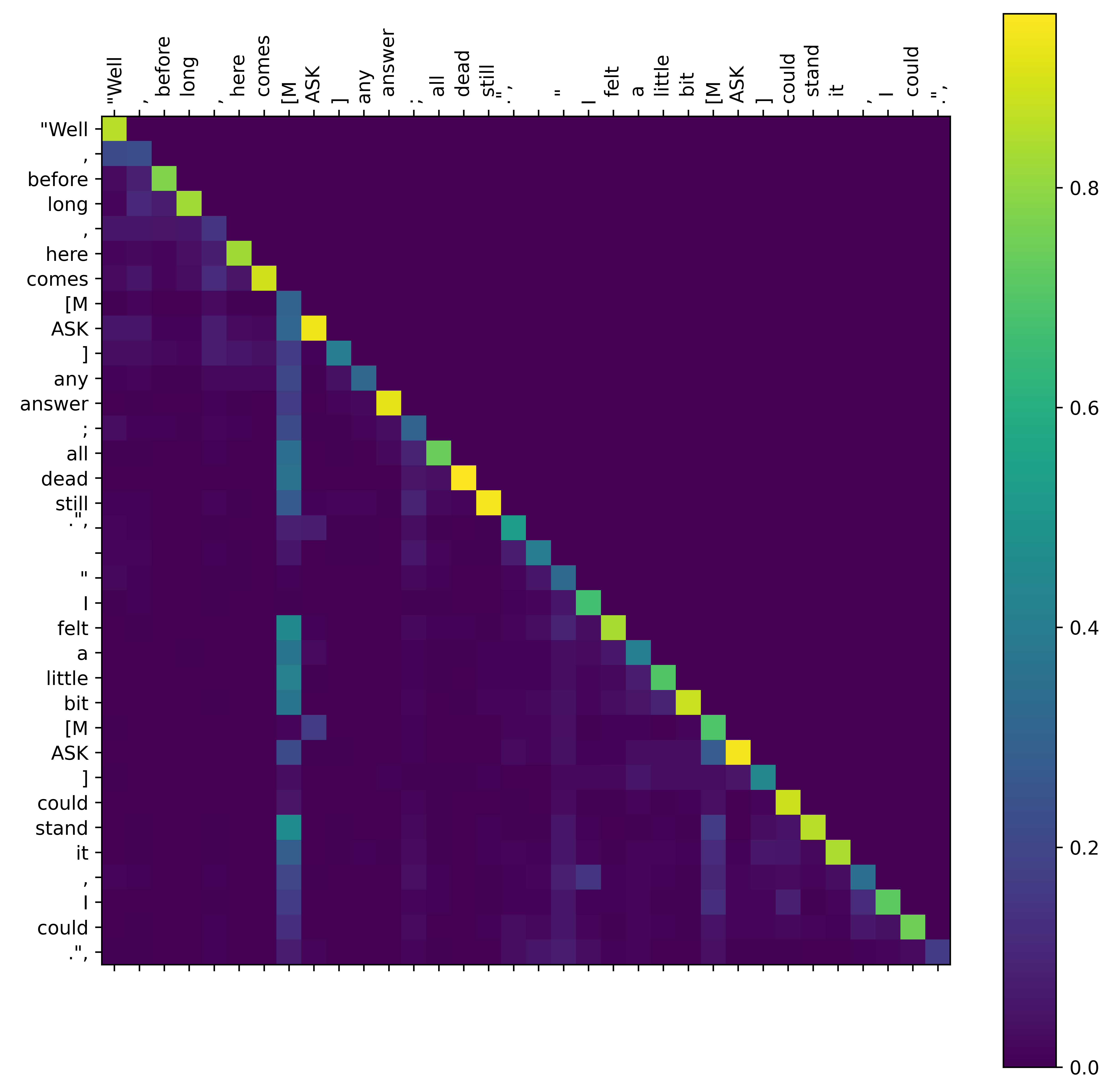}
    \caption{Score distribution of attention heads before fine-tuning.}
    \label{atten1}
\end{figure}
\begin{figure}[h]
    \centering
    \includegraphics[width=0.47\textwidth]{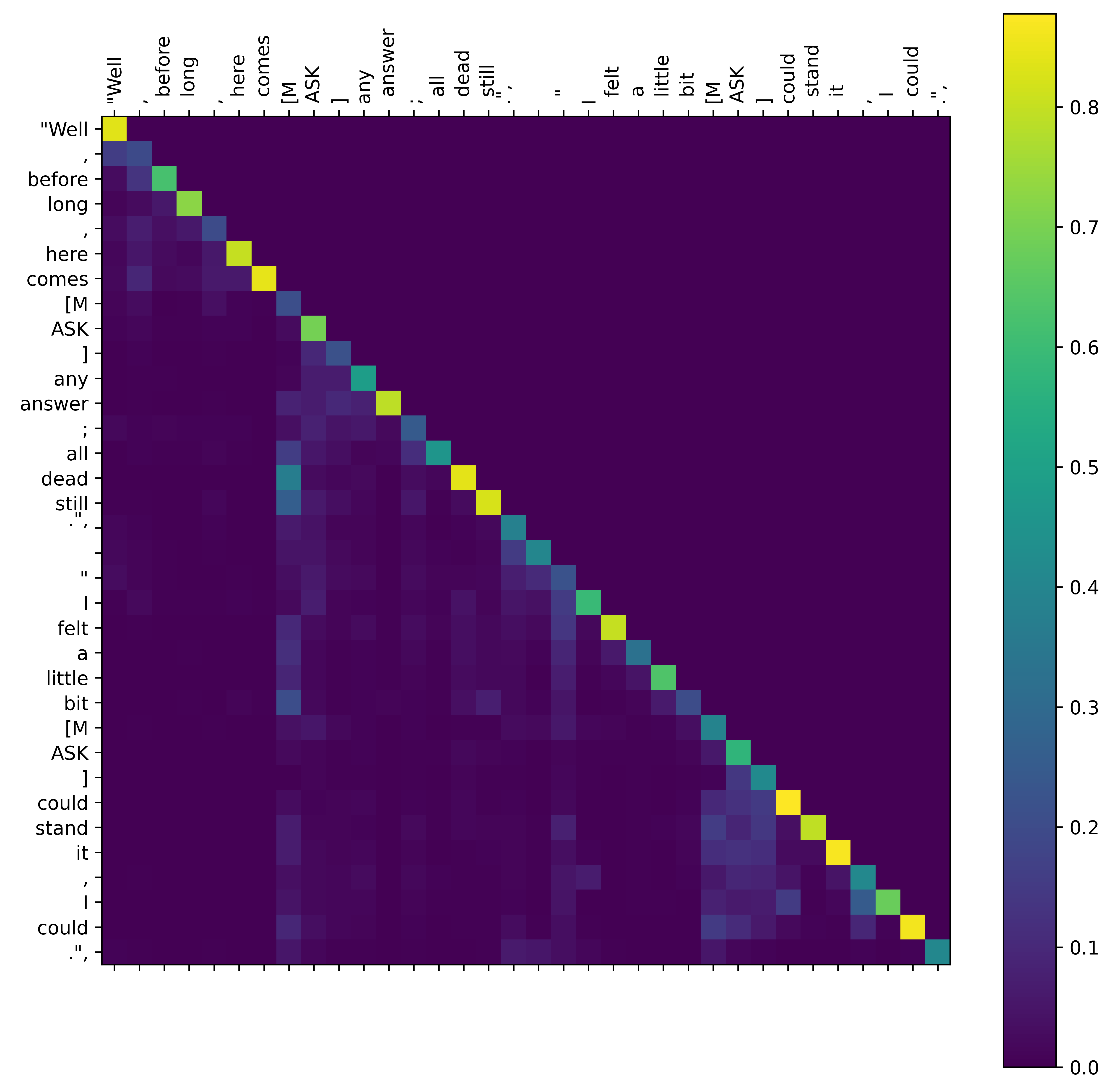}
    \caption{Score distribution of attention heads after fine-tuning.}
    \label{atten2}
\end{figure}

\begin{figure}[h]
    \centering
    \includegraphics[width=0.5\textwidth]{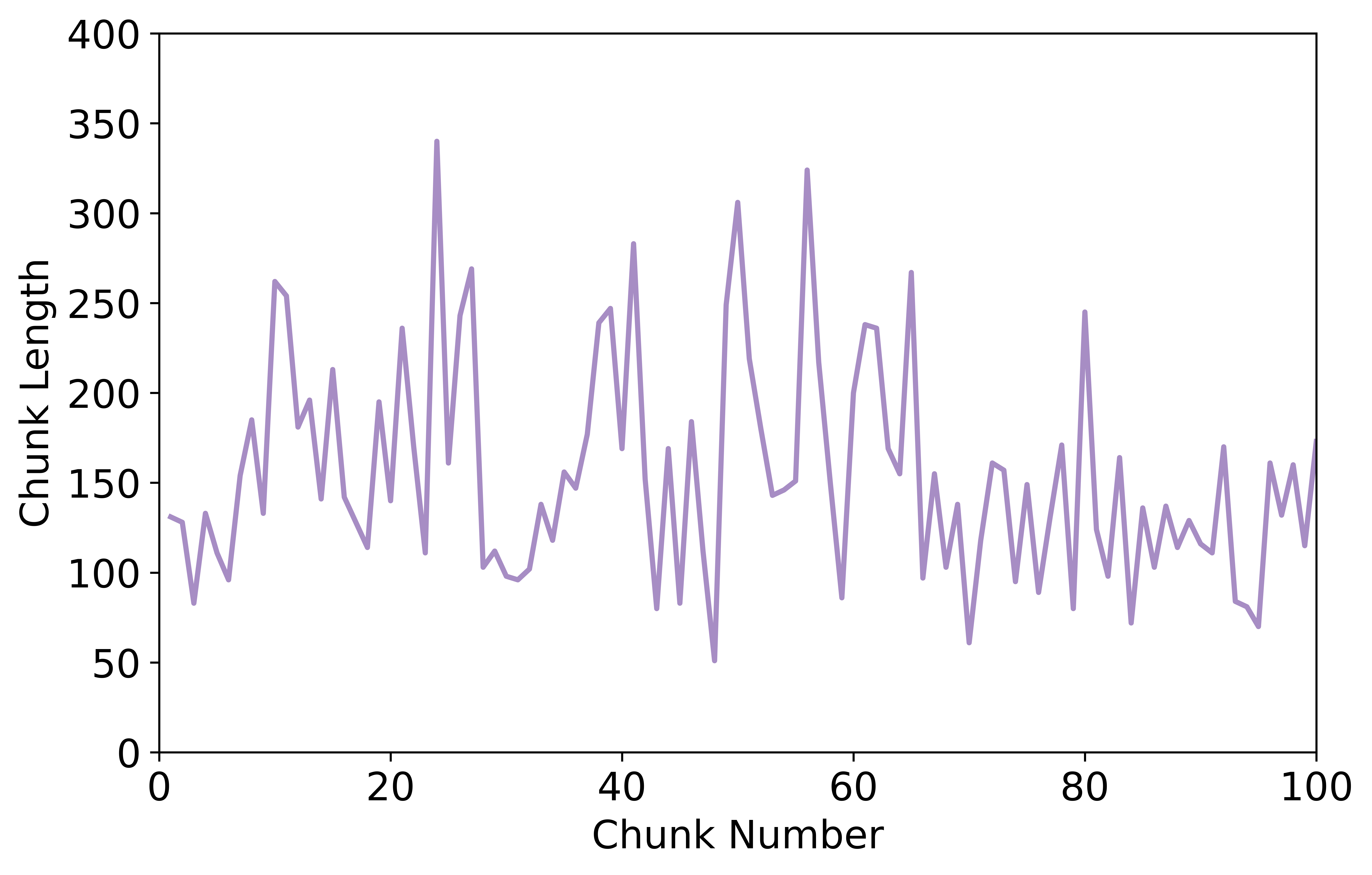}
    \caption{Granularity distribution of text chunks generated by GPT-4o on the CRUD benchmark.}
    \label{Chunk_granularity}
\end{figure}

During the model training phase, the paper designs a multi-granularity aware router capable of dynamically selecting the most suitable chunker for documents requiring different chunk lengths, as depicted in Figure \ref{Chunk_granularity}. Furthermore, we train a meta-chunker using a full fine-tuning strategy to generate structured chunking regular expressions. To address potential hallucination issues, we introduce an edit distance recovery algorithm, ensuring the accuracy of chunking. The entire process not only improves data quality and chunking precision but also maintains efficient computational capabilities through the sparse activation mechanism, replacing chunk generation with chunking rules generation, and utilizing smaller models in place of larger ones.

\renewcommand{\arraystretch}{1.4} 
\setlength{\extrarowheight}{1pt} 
\begin{table*}[t]
\centering
\resizebox{\textwidth}{!}{%
\begin{tabular}{lccc|ccc|c|c}
\toprule
\multirow{2}{*}{\textbf{Chunking Methods}} & \multicolumn{3}{c}{\textbf{CRUD (Single-hop)}} & \multicolumn{3}{c}{\textbf{CRUD (Two-hop)}}  & \multicolumn{1}{c}{\textbf{DuReader}} & \multicolumn{1}{c}{\textbf{WebCPM}} \\
 &\textbf{BLEU-1} & \textbf{BLEU-Avg} & \textbf{ROUGE-L} & \textbf{BLEU-1} & \textbf{BLEU-Avg} & \textbf{ROUGE-L}   & \textbf{F1}  &  \textbf{ROUGE-L} \\
\midrule
Original & $0.3515_{-0.002}$ & $0.2548_{-0.0014}$ & $0.4213^{+0.0005}_{-0.0014}$ & $0.2322^{+0.003}$ & $0.1133^{+0.0023}$ & $0.2613^{+0.0026}$ & $0.2030^{+0.0029}$ & $0.2642^{+0.0007}_{-0.004}$ \\
Llama\_index & $0.3620_{-0.0109}$ & $0.2682_{-0.0089}$ & $0.4326_{-0.0115}$ & $0.2315^{+0.0005}$ & $0.1133^{+0.0016}$ & $0.2585^{+0.0026}$ & $0.2220_{-0.0076}$ & $0.2630^{+0.0034}$ \\
Semantic Chunking & $0.3382^{+0.0007}_{-0.0012}$ & $0.2462^{+0.0015}_{-0.0006}$ & $0.4131^{+0.0005}_{-0.0012}$ & $0.2223^{+0.0024}_{-0.0003}$ & $0.1075_{-0.0012}$ & $0.2507^{+0.0004}$ & $0.2157_{-0.0061}$ & $0.2691_{-0.0013}$ \\
LumberChunker & $0.3456_{-0.0052}$ & $0.2542_{-0.0037}$ & $0.4160_{-0.0037}$ & $0.2204^{+0.0003}_{-0.0006}$ & $0.1083^{+0.0008}_{-0.0004}$ & $0.2521^{+0.0006}_{-0.001}$ & $0.2178^{+0.0033}$ & $0.2730^{+0.0024}$ \\
\addlinespace[2pt] 
\cdashline{1-9} 
Qwen2.5-14B & $0.3650_{-0.0035}$ & $0.2679^{+0.0004}_{-0.0015}$ & $0.4351^{+0.0001}_{-0.0029}$ & $0.2304^{+0.0001}$ & $0.1129^{+0.0007}$ & $0.2587^{+0.0023}$ & $0.2271^{+0.003}$ & $0.2691^{+0.0011}$ \\
Qwen2.5-72B & $0.3721^{+0.0019}$ & $0.2743^{+0.0015}$ & $0.4405^{+0.0023}$ & $0.2382_{-0.0053}$ & $0.1185_{-0.004}$ & $0.2677_{-0.0053}$ & $0.2284^{+0.003}$ & $0.2693^{+0.0017}$ \\
\addlinespace[2pt] 
\cdashline{1-9} 
Meta-chunker-1.5B & $0.3754_{-0.0028}$ & $0.2760^{+0.0003}_{-0.0021}$ & $0.4445^{+0.0002}_{-0.0025}$ & $0.2354^{+0.0018}_{-0.0007}$ & $0.1155^{+0.0022}_{-0.0001}$ & $0.2641^{+0.0021}$ & $0.2387^{+0.0036}_{-0.002}$ & $0.2745^{+0.0023}_{-0.0012}$ \\
\bottomrule
\end{tabular}%
}
\caption{Confidence intervals of main results determined through three independent experiments.}
\label{main-performance2}
\end{table*}

On the other hand, within the current research context, traditional chunking methods exhibit several limitations. Specifically, existing chunking techniques typically require sentence segmentation prior to chunking sentences. This approach presents several evident issues:

\begin{itemize}
    \item Sentence boundaries in natural language are not always clear and unambiguous, particularly in texts containing extensive colloquial or non-standard usage, where automatic sentence segmentation may introduce errors.
    \item Simple sentence segmentation methods fail to accurately process texts with complex structures, nested sentences, or long sentences. This can result in the omission or misinterpretation of crucial information.
    \item Texts often contain structured information such as tables and code, and chunking by sentence may disrupt these structures, making it difficult to correctly understand and process the information.
\end{itemize}

Therefore, it is necessary to explore a novel chunking method to overcome the limitations of the segment-then-chunk approach. Driven by this motivation and necessity, we propose an end-to-end solution for text chunking using SLMs directly. 

Notably, further discoveries are made when analyzing the attention score heatmaps of models within the MoC framework, as illustrated in Figures \ref{atten1} and \ref{atten2}. Before fine-tuning, the attention heads assigned high scores to special tokens, but these scores decreased after fine-tuning, while the attention scores for characters at the beginning and end of text chunks increased. This indicates that before fine-tuning, the model's attention allocation has not yet fully adapted to the specific task requirements. Some attention heads excessively focus on special tokens in the initial stage, which hinders the normal completion of the task. During the fine-tuning process, the model redistributes and adjusts its attention based on the data and objectives of the specific task. Attention heads that are initially overly focused on a certain token reduce their scores and shift their attention more towards information that is more relevant to the task. Consequently, through fine-tuning, the model's understanding and mastery of the task improve, enabling it to more accurately identify and focus on the information that is truly important for task completion. This further demonstrates the effectiveness of our approach. 

To ensure statistical robustness, we conduct three independent experiments for each method to determine the boundaries of confidence intervals, with some data showing a tendency towards one side. The specific experimental results are presented in Table \ref{main-performance2}.

\subsection{Relationship Between BC, CS, and the Performance of the RAG System}
\label{Relationship3}
To more accurately demonstrate the correlation between BC, CS, and the performance of the RAG system, we add the calculation of Pearson correlation coefficients based on the existing experiments in the paper. The Pearson correlation coefficient is a statistical measure that assesses the degree of linear correlation between two variables, with a value range of [-1, 1]. A positive correlation coefficient indicates a positive linear relationship between the two variables; a negative correlation coefficient indicates a negative linear relationship; and a correlation coefficient of 0 indicates no correlation between the two variables. If the absolute value of the correlation coefficient is greater than 0.5, it suggests a relatively evident linear relationship between the variables. 

We select the evaluation results of the Qwen2.5-7B, as shown in the table \ref{tab:Relationship}. After calculation, the correlation coefficients between BC, $\text{CS}_c$, $\text{CS}_i$, and ROUGE-L are found to be 0.8776, -0.7453, and -0.6663, respectively. This indicates that there is indeed a noticeable linear relationship between BC, CS, and the performance of the RAG system, further validating our discussion in the paper: BC represents boundary clarity, which is preferable when higher; $\text{CS}_c$ denotes chunk stickiness utilizing a complete graph, and $\text{CS}_i$ indicates chunk stickiness employing a incomplete graph, both of which are favorable when lower.

\renewcommand{\arraystretch}{1.4} 
\setlength{\extrarowheight}{1pt} 
\begin{table}[t]
\centering
\resizebox{0.47\textwidth}{!}{%
\begin{tabular}{lcccc}
\toprule
 \textbf{Chunking Methods} & \textbf{ROUGE-L} & \textbf{BC} & \textbf{$\text{CS}_c$}& \textbf{$\text{CS}_i$}   \\
\midrule
Original & 0.4213 & 0.8049 & 2.421 & 1.898  \\
Llama\_index & 0.4326 & 0.8455 & 2.250  & 1.483  \\
Semantic Chunking & 0.4131 & 0.8140 & 2.325 & 1.650 \\
Qwen2.5-14B & 0.4351 & 0.8641 & 2.125 & 1.438  \\
\bottomrule
\end{tabular}%
}
\caption{Performance of different chunking methods under Qwen2.5-7B.}
\label{tab:Relationship}
\end{table}

\subsection{Prompt utilized in Chunking}
When preparing datasets using GPT-4o and generating chunking rules with MoC, prompts are necessary, as illustrated in Tables \ref{zztab:7} and \ref{zztab:8}. The design and implementation of these prompts are crucial, as they directly influence the quality and characteristics of the resulting datasets and chunking rules.

\subsection{Details of MoC Training}
\begin{figure}[h]
    \centering
    \includegraphics[width=0.5\textwidth]{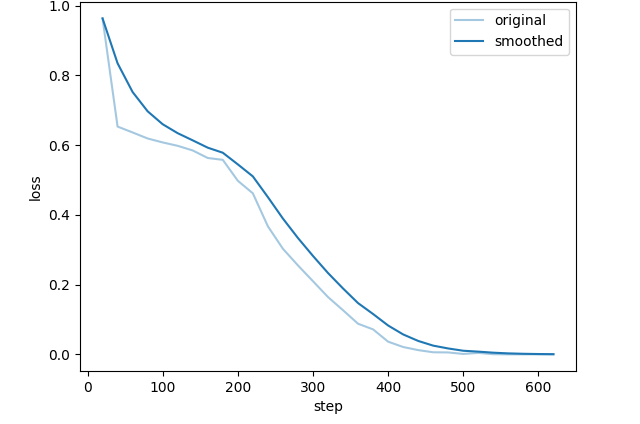}
    \caption{Trend of loss change during router training.}
    \label{zzm1}
\end{figure}

\begin{figure}[h]
    \centering
    \includegraphics[width=0.5\textwidth]{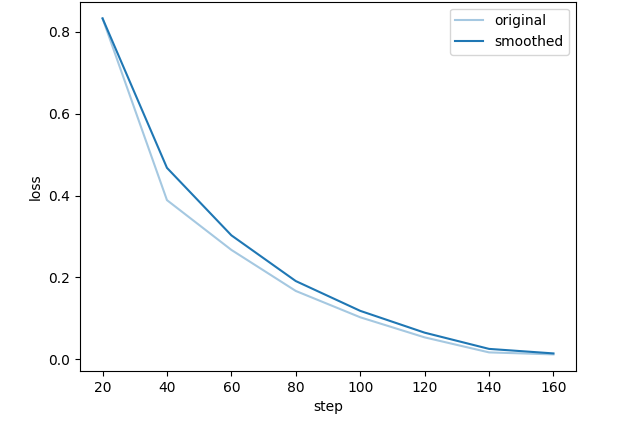}
    \caption{Trend of loss change during meta-chunker training with granularity range [0,120].}
    \label{zzm2}
\end{figure}

\begin{figure}[h]
    \centering
    \includegraphics[width=0.5\textwidth]{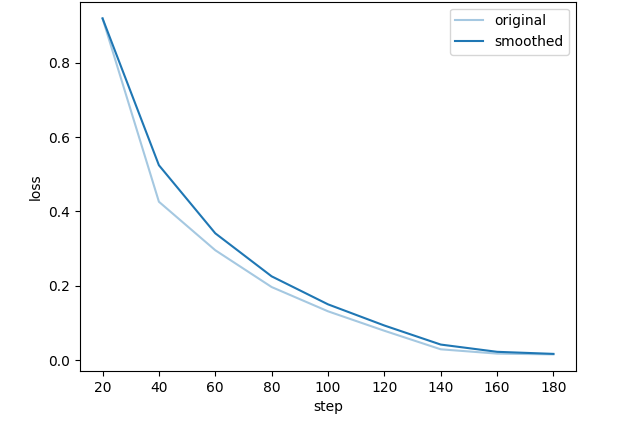}
    \caption{Trend of loss change during meta-chunker training with granularity range (120,150].}
    \label{zzm3}
\end{figure}

\begin{figure}[h]
    \centering
    \includegraphics[width=0.5\textwidth]{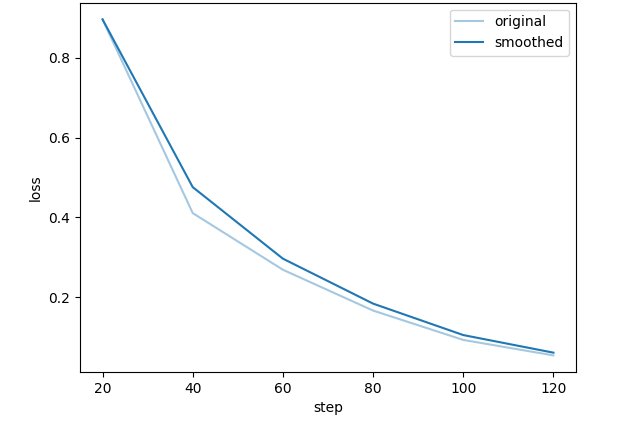}
    \caption{Trend of loss change during meta-chunker training with granularity range (150,180].}
    \label{zzm4}
\end{figure}

\begin{figure}[h]
    \centering
    \includegraphics[width=0.5\textwidth]{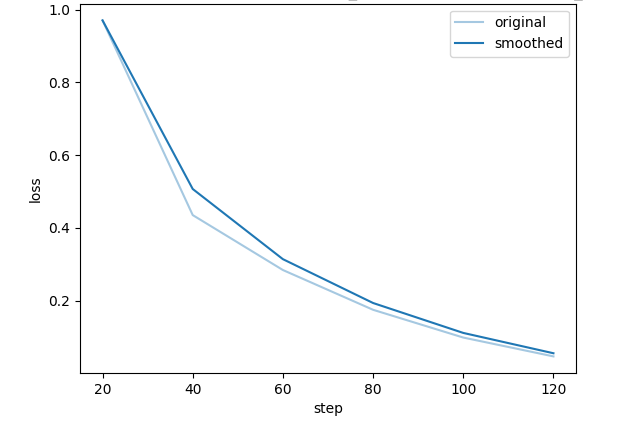}
    \caption{Trend of loss change during meta-chunker training with granularity range $(180,+\infty)$.}
    \label{zzm5}
\end{figure}

During the model training phase, we adopt specific parameter configurations. Specifically, the training batch size per device is set to 3, and model parameters are updated every 16 steps through the gradient accumulation strategy. The learning rate is set to $1.0e-5$ to achieve fine-grained adjustment of weights. The model underwent a total of 3 epochs of training. Additionally, we employ a cosine annealing learning rate scheduling strategy and set a warmup ratio of 0.1 to facilitate model convergence. The variations in training loss are recorded, namely, Figure \ref{zzm1} showcases the training loss of the router, while Figure \ref{zzm2}-\ref{zzm5} individually depict the training losses of chunking experts in different intervals. The bf16 format is enabled during training to balance memory consumption and training speed. This training is conducted on two NVIDIA A800 80G graphics cards, ensuring efficient computing capabilities.

\begin{figure*}[t]
    \centering
    \includegraphics[width=\textwidth]{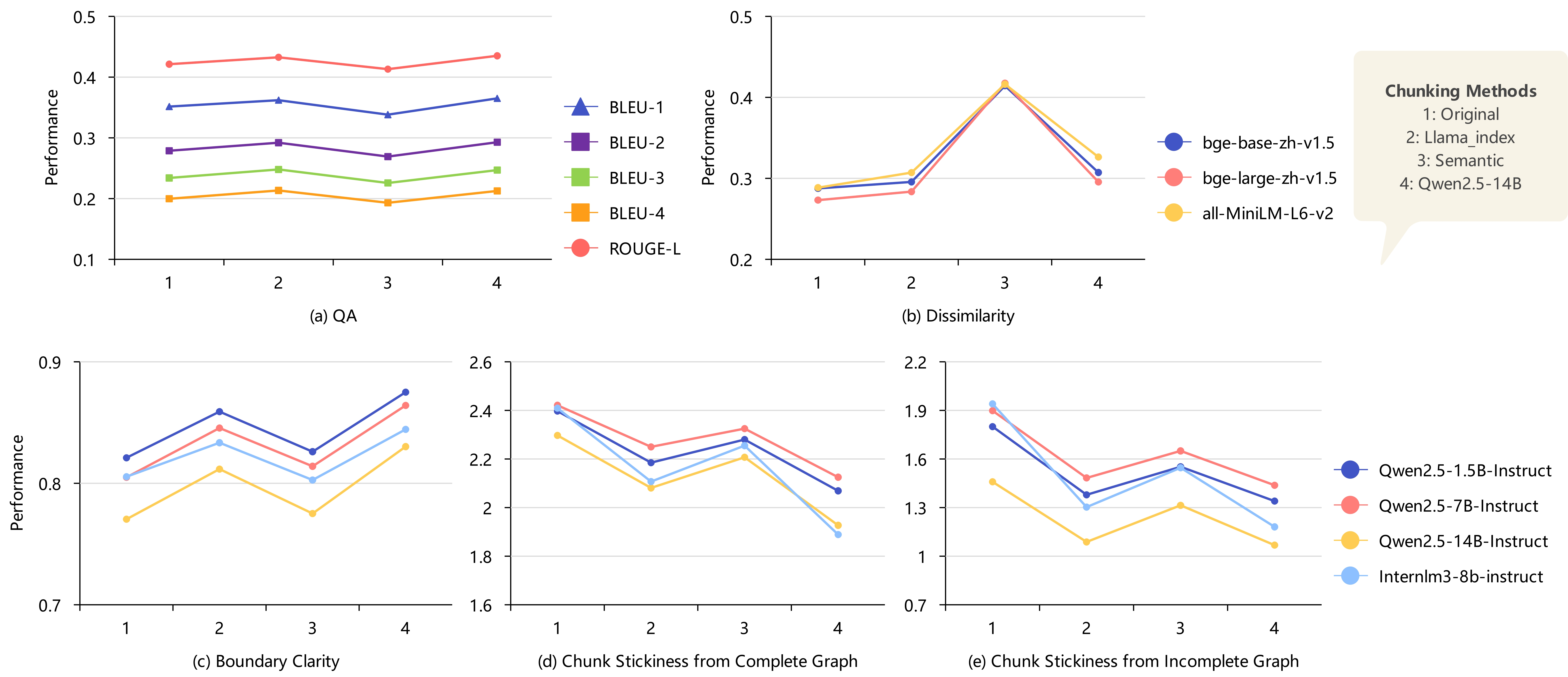}
    \caption{Trends in evaluating chunking performance using different metrics.}
    \label{picExploring Chunking}
\end{figure*}


\renewcommand{\arraystretch}{1.4} 
\setlength{\extrarowheight}{1pt} 
\begin{table}[h]
\centering
\resizebox{0.47\textwidth}{!}{%
\begin{tabular}{p{0.47\textwidth}}
\toprule
\textbf{Chunking Prompt} \\
\hline
This is a text chunking task, and you are an expert in text segmentation, responsible for dividing the given text into text chunks. You must adhere to the following four conditions:\\
1. Segment the text based on its logical and semantic structure, ensuring each text chunk expresses a complete logical thought.\\
2. Avoid making the text chunks too short, balancing the recognition of content transitions with appropriate chunk length.\\
3. Do not alter the original vocabulary or content of the text.\\
4. Do not add any new words or symbols.\\
If you understand, please segment the following text into text chunks, with each chunk enclosed using <chunk> and </chunk>. Output the complete set of segmented chunks without omissions.\\\\

Document content: [Text to be segmented]\\

The segmented text chunks are:\\
\bottomrule
\end{tabular}%
}
\caption{Prompt for direct chunking of GPT-4o.}
\label{zztab:7}
\end{table}

\renewcommand{\arraystretch}{1.4} 
\setlength{\extrarowheight}{1pt} 
\begin{table}[h]
\centering
\resizebox{0.47\textwidth}{!}{%
\begin{tabular}{p{0.47\textwidth}}
\toprule
\textbf{Chunking Prompt} \\
\hline
This is a text chunking task. As an expert in text segmentation, you are responsible for segmenting the given text into text chunks. You must adhere to the following four conditions:\\
1. Combine several consecutive sentences with related content into text chunks, ensuring that each text chunk has a complete logical expression.\\
2. Avoid making the text chunks too short, and strike a good balance between recognizing content transitions and chunk length.\\
3. The output of the chunking result should be in a list format, where each element represents a text chunk in the document.\\
4. Each text chunk in the output should consist of the first few characters of the text chunk, followed by "[MASK]" to replace the intermediate content, and end with the last few characters of the text chunk. The output format is as follows:\\
\text{[}\\
\quad "First few characters of text chunk [MASK] Last few characters of text chunk",\\
\quad ......\\
\text{]}\\
If you understand, please segment the following text into text chunks and output them in the required list format.\\
\\
Document content: [Text to be segmented]\\
\bottomrule
\end{tabular}%
}
\caption{Prompt for chunking of MoC.}
\label{zztab:8}
\end{table}

\end{document}